# Unify Change Point Detection and Segment Classification in a Regression Task for Transportation Mode Identification

**Rongsong Li**
Department of Automation, Tsinghua University
Beijing, China 100084
Email: lirs17@tsinghua.org.cn

**Xin Pei**
Department of Automation, Tsinghua University
Beijing, China 100084
Email: peixin@mail.tsinghua.edu.cn

**ABSTRACT**

Identifying travelers' transportation modes is important in transportation science and location-based services. It's appealing for researchers to leverage GPS trajectory data to infer transportation modes with the popularity of GPS-enabled devices, e.g., smart phones. Existing studies frame this problem as classification task. The dominant two-stage studies divide the trip into single-one mode segments first and then categorize these segments. The over segmentation strategy and inevitable error propagation bring difficulties to classification stage and make optimizing the whole system hard. The recent one-stage works throw out trajectory segmentation entirely to avoid these by directly conducting point-wise classification for the trip, whereas leaving predictions dis-continuous. To solve above-mentioned problems, inspired by YOLO and SSD in object detection, we propose to reframe change point detection and segment classification as a unified regression task instead of the existing classification task. We directly regress coordinates of change points and classify associated segments. In this way, our method divides the trip into segments under a supervised manner and leverage more contextual information, obtaining predictions with high accuracy and continuity. Two frameworks, TrajYOLO and TrajSSD, are proposed to solve the regression task and various feature extraction backbones are exploited. Exhaustive experiments on GeoLife dataset show that the proposed method has competitive overall identification accuracy of 0.853 when distinguishing five modes: walk, bike, bus, car, train. As for change point detection, our method increases precision at the cost of drop in recall. All codes are available at https://github.com/RadetzkyLi/TrajYOLO-SSD.

# INTRODUCTION

Identifying travelers' transportation or travel modes is important in many fields. With such knowledge, we can optimize transportation planning and city planning so as to reduce construction cost, traffic congestion and air pollution in transportation science; achieve accurate advertising in location-based services; mine human daily activity patterns in human geography (*1,2*). Traditionally, the knowledge of travel modes is obtained by household questionnaire and telephone interviews. However, such methods are labor-intensive, time-consuming, expensive, with low responsive rate and inaccurate information. Thus, an automatic and cost-effective technology is needed.

With the popularity of GPS-enabled devices, especially smart phones, researchers are increasingly turning to leverage GPS data to infer transportation modes. GPS-enabled devices can automatically record moving objects' spatial-temporal information nearly all the time. Due to high market penetration rate of GPS-enabled devices, it becomes possible to collect massive trajectories of people. Typically, such device can record its carrier's position at a given time interval. The chronological records from a device constitute the trajectory or track.

Most of the inference models on detecting travel modes using GPS trajectory data follow a two-stage paradigm proposed by Zheng et al. (*3-5*). In the first stage, a trip is divided into segments with only one transportation mode by detecting change points, i.e., position where transportation mode changes. The process relies on expert knowledge, heuristics and assumptions, e.g., walk is intermediate between other two travel modes (*5*). In the second stage, we extract features of the segment and then categorize it into specific transportation mode. Various classifiers have been exploited, e.g., Multi-layer Perceptron (MLP), Support Vector Machine (SVM), Decision Tree (DT), Random Forest (RF) (*1-5*) and deep learning algorithms (*6*). In addition, a recent study (*7*) tried to directly predict transportation mode of each GPS point in a trip without partitioning a trip into segments, which was called the one-stage method.

However, for two-stage methods, the change point detection is unsupervised and dependent heavily on expert knowledge which is time-lagged for modern traffic changes. The inevitable segmentation error in first stage would propagate into the classification stage and amplify. Also, the over segmentation strategy, i.e., dividing the trip into very short segments to recall change points as possible (*2*), limits the utilization of broader contextual information and thus brings difficulty to classification. As confirmed by Li et al. (*7*) by experiments, segmentation error has serious effect on following classification, however, such influence was ignored by most of two-stage studies because they concentrated only on improving classifiers and divided a trip into segments by annotations in all experiments. Hence, the overall identification is unknow if deployed in real application, i.e., inferring transportation modes from unlabeled GPS trajectory. Owing to independence of the two stages, it's challenging to adjust related modules when optimizing the whole system.

As for one-stage methods, they are easy to be optimized to improve the overall identification performance, but would produce dis-continuous predictions, i.e., too many change points, which is unrealistic. Both of current two-stage and one-stage methods frame transportation modes using GPS trajectory data as a classification task.

To solve problems mentioned-above, we draw inspiration from YOLO (You Only Look Once) (*8*) and SSD (Singe Shot Detector) (*9*), classic networks in computer vision, which reframe object detection as a regression task instead of the previous classification task by predicting bounding boxes and class probabilities simultaneously, Building on this idea, we propose to reframe change point detection and segment classification as a single regression problem, straight from GPS trajectory to coordinates of change points and class probabilities of segments. Under this way, compared to existing two-stage methods, for the first time, we can bring supervisory information for change point detection, incorporate more contextual information, and optimize the entire process effectively. Compared to one-stage methods, this modeling naturally produces continuous predictions. To solve the regression task, two frameworks, TrajYOLO and TrajSSD, are proposed and various feature extraction backbones are exploited.

We summarize our main contributions as follows:

(1) We propose to frame change point detection and segment classification as a single regression problem for transportation mode identification based-on GPS trajectory data, leading to end-to-end training and inference, which is convenient for optimizing the whole system and produce predictions with high accuracy and continuity. Such formula reframes change point detection from non-supervision to supervision.

(2) We construct two frameworks, TrajYOLO and TrajSSD, to solve the reframed regression problem, in which various feature extractors and network designs are exploited to mine more detailed information and locate change points more accurate. In addition, both frameworks receive trips with various lengths.

(3) We conduct extensive experiments on GeoLife dataset to demonstrate the superior performance of the proposed method. Empirical studies also reveal that our model is light-weighted and fast.

The rest of this paper is organized as follows. Section II reviews literatures related to change point detection and transportation mode identification. Section III describes the used dataset and the pre-processing steps. Section IV presents the proposed regression-based modeling method and details the implementation. Section V demonstrates the experimental results and section VI concludes this paper.

**LITERATURE REVIEW**

Transportation mode identification is fundamental in fields of transportation science, location-based services and human geography (*2*). There are various data sources for this purpose, including GPS, Global System for Mobile communications (GSM), accelerometer, Geographical Information System (GIS) and so on (*11*). We review most related studies which use GPS data, including two-stage and one-stage methods.

In two stage studies, the first segmentation stage, i.e., dividing a trip into single-one mode segments, is implemented by detecting change points or transition points. Existing segmentation algorithms can be divided into 3 categories: walk segment-based, sliding window and similarity-based and inflection points-based. Zheng et al. (*3-5*) proposed the walk segment-based method, they assumed that when people change their transportation mode, they must take walk as transition. By identifying walk points via threshold of velocity and acceleration, a trip is divided into walk segments and non-walk segments. Then, short segments and uncertain segments are merged. Efforts were put to refine the merge process, e.g., Zhu et al. (*12*) and Li et al. (*13*) defined a set of rules considering real world constraints respectively, Chen et al. (*14*) introduced density-based clustering method. For sliding window and similarity-based methods, the easiest way is to slide uniform time window over the trip to generate sub-trips, which is short enough to be segments. Then, one can define and calculate the similarity of neighboring sub-trips and merge them (*15*). For the last one, Dabiri et al. (*16*) created time-series signals (e.g., velocity and acceleration) from GPS trajectories and presumed the transition point correspond to inflection point of the time-series, which was searched by Pruned Exact Linear Time (PELT) method. Nevertheless, existing methods are unsupervised and rely heavily on expert knowledge.

In the second classification stage, both traditional machine learning and deep learning algorithms have been explored. For the former, hand-crafted features of the segment are created first and then fed into a classifier. Commonly used hand-crafted features encompass statistics (e.g., maximum, minimum, mean, variance) of distance, velocity, acceleration, jerk and head change (*17*). All kinds of classifiers were put into use, including K-nearest Neighbors (KNN), Bayesian Networks, SVM, DT, RF and conditional random field (*3-5,18,19*). Among them, SVM and RF outperform others.

Extracting features manually depends on expertise, therefore, researchers started to take advantages of deep learning excelling in feature extraction and representation. The challenge is to construct proper inputs because neural networks usually accept size-fixed and structured data, e.g., image. Dabiri and Heaslip (*6*) made the significant step, by creating a four-channel structure for the segment in which four channels are speed, acceleration, jerk and bearing rate respectively. Such representation combined with CNN suppressed traditional methods. Afterwards, many researchers expended this work on the same GeoLife dataset: Yu (*20*) concatenated high level features from Long Short-Term Memory (LSTM) and frequency features from discrete wavelet transformation; Li et al. (*21*) utilized Generative Adversarial Networks (GAN) to enhance dataset; Dabiri et al. (*16*) proposed SEmi-Supervised Convolutional Autoencoder (SECA) to leverage unlabeled trajectories. In addition, Jia et al. (*22*) made use of bayes neural networks for relieving overfitting.

Nevertheless, current two-stage studies are hard to optimize, i.e., figuring out whether unsatisfactory predictions result from poor segmentation or vulnerable classification. To tackle this, Li et al. (*7*) proposed a one-stage framework in which trajectory segmentation is throw out entirely by directly conducting pointwise classification. Nonetheless, predictions from one-stage methods are fatally dis-continuous, needing extra post-processing. To handle these drawbacks, we propose to reframe change point detection and segment classification as a unified regression problem in the following sections.

**DATA DESCRIPTION AND PRE-PROCESSING**

In this section, we first introduce the widely used GPS trajectory dataset from GeoLife project from Microsoft Research Asia and define several terms used in this paper. Then we detail the procedure of data pre-processing.

**Dataset**

The GeoLife GPS trajectory dataset was collected by 182 users in a period from 2007 to 2012 (*5*). Majority of the data was created in cities in China, especially Beijing, with a dense sampling rate, i.e., more than 91.5% of trajectories were recorded every 1-5 seconds or 5-10 meters. 69 users annotated their trajectories with the aid of trajectories visualization on the map. There are various labels (e.g., walk, bike, airplane) in raw annotations but only land transportation modes (i.e., walk, bike, car, taxi, bus, subway and train) remains after matching annotation files with trajectory files using time-stamp. As suggested in the dataset user guide, we assign the label of taxi and car as car due to their similar motion characteristics. In addition, because there is no GPS signal underground and reported subway is in fact light rail connected to subway, we refer subway and train as train. Finally, the five transportation modes need identifying: walk, bike, bus, car and train.

It's worth noting that Li et al. (*7*) found annotations were not totally consistent with actual map, e.g., trajectories labeled as subway appeared in highway in suburbs of Beijing, and corrected annotations manually related to subway and train. The difference between original and corrected dataset is listed in **Figure 1**. The main change is that the amount of subway dropped by nearly 40%. The corrected dataset from their code repository is used.

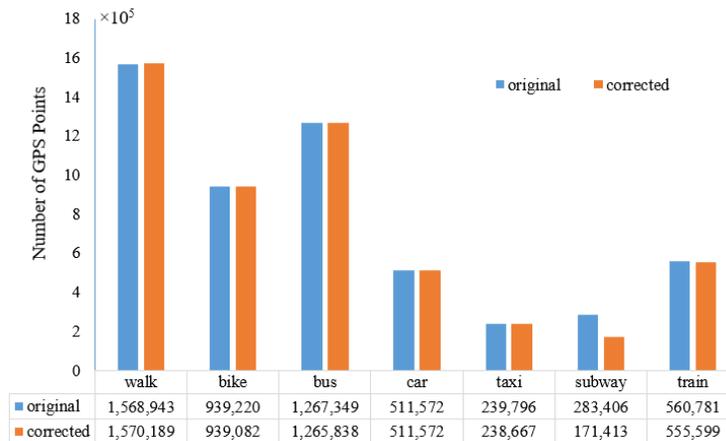

**Figure 1 Description of original and corrected dataset**

**Preliminaries**

For clarification, the following terms should be clearly defined: GPS point, GPS trajectory, trip, segment and change point.

GPS *point*, known as GPS record, identifies the geographical location of a moving object at some moment. Usually, it is denoted by a tuple *P = (t, lat, lng)* where *t*, *lat* and *lng* represent time-stamp, latitude and longitude respectively.

GPS *trajectory*, also known as GPS *track*, is a chronological collection of GPS records over a certain period. The length of trajectory denotes the number of GPS points in the trajectory, and the distance of trajectory denotes the accumulated geodesic distance during the trajectory.

A *trip* is the trajectory recorded during a purposeful movement such as from home to company. Various transportation modes can be used in a trip, e.g., we walk from home to bus station, then go to bus station near company by bus, and finally walk to company. The trajectory is partitioned into trips if time interval between two consecutive GPS points exceeds a pre-defined threshold.

A *segment* is the sub-trip during which only one transportation mode is utilized. For instance, the trip from home to company mentioned-above consists of two walk segments and one bus segment.

A *change point* or *transition point*, denoted as CP, is defined as the position where traveler change their transportation mode. Thus, a trip can contain zero, one or more change points. Clearly, n change points mean n+1 segments in a trip. The illustration of trip, segment and change point is in **Figure 2**.

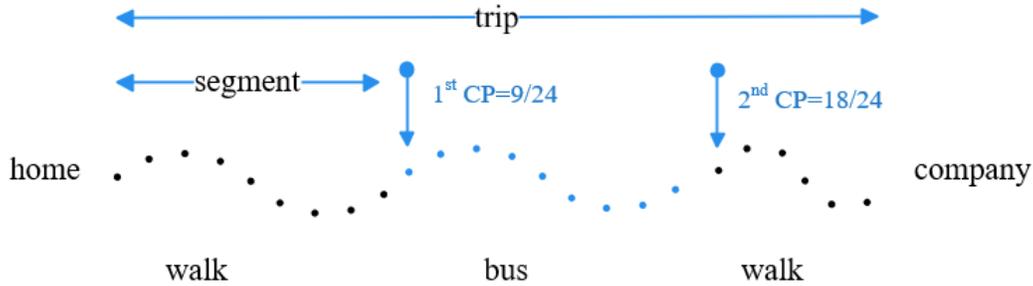

Figure 2 Trip, segment and change point (CP)

**Data Pre-processing Procedure**

The raw GPS data contains only geographical coordinates and time-stamps, which is intractable to infer transportation modes. Generally, researchers would develop some motion characteristics to fuse geographical information and expertise. The following motion characteristics built by Dabiri and Heaslip (*6*) have been proven to be discernable and effective by many studies (*20-21*): relative distance, speed (S), acceleration (A), jerk (J) and bearing rate. Among them, combination of speed, acceleration and jerk worked best (*7*), which we followed in this paper. After dividing the trajectory into trips by threshold of 20 minutes (*5*), we calculate these motion features for each trip as **Equation 1-3**:

$$S_{P_i} = \frac{Vincenty(P_i, P_{i+1})}{\Delta t} \quad (1)$$

$$A_{P_i} = \frac{S_{P_{i+1}} - S_{P_i}}{\Delta t} \tag{2}$$

$$J_{P_i} = \frac{A_{P_{i+1}} - A_{P_i}}{\Delta t} \tag{3}$$

where $\Delta t, S_{P_i}, A_{P_i}, J_{P_i}$ denote time difference, speed, acceleration and jerk between two consecutive GPS points $P_i$ and $P_{i+1}$, respectively. The geodesic distance between two GPS points is calculated using Vincenty formula (*23*). Values of these characteristics of the first GPS point are set to zero, indicating we start a trip from stillness.

To remove random errors and reduce noises, we apply the five-spot triple smoothing method to smooth characteristics for each trip. This smoothing method is a variant of Savitzky-Golay filter which is strong in maintaining the shape and pattern of signal (*7*).

Before processing steps stated above, however, we need to take the following steps to remove wrong and isolated GPS records resulting from occlusion of skyscrapers, tunnel and atmosphere disturbances (*6*).

- The GPS point with time-stamp less than its previous one is identified as erroneous and discarded.
- Calculate the speed and acceleration of remaining GPS points according to **Equation 1** and **Equation 2**.
- The GPS point whose speed or acceleration exceeds certain threshold is identified as outlier and discarded. This paper focus on land transportation modes, thus speed and acceleration threshold are selected as 80 m/s and 10 m/s$^2$, belonging to train and car respectively.

Finally, we constrain the length of the trip between $N_{\min}$ and $N_{\max}$ as in Li et al. (*7*) which also received trips as inputs by dividing long trips into non-overlapped trips of length not greater than $N_{\max}$ and discarding short trips of length less than $N_{\min}$. The purpose is to let the trip contain sufficient but not redundant information. We set $N_{\min} = 20$ and $N_{\max} = 400$ which worked best in Li et al. (*7*).

After all pre-processing steps, we obtain trips of length between 20 and 400, represented by $N \times 3$ matrices, where $N$ denotes the number of GPS points in the trip and 3 denotes three motion characteristics (i.e., speed, acceleration and jerk).

**METHODOLOGY**

In this section, we first demonstrate the proposed unified regression task and then introduce two frameworks to solve this task. Finally, matters needing attention during training are detailed.

**The Proposed Unified Regression Task**

As depicted in **Figure 3**, the proposed regression task unifies change point detection and segment classification in a problem. We can calculate coordinates of change points and predict class of associated segments concurrently. Compared to two-stage framework of classification task, our model simplifies the optimization process for improving the overall system performance and detects change points under supervised instead of unsupervised manner. Compared to one-stage framework of classification task, our model can produce continuous predictions naturally. Besides, the following research questions (RQ) need addressing.

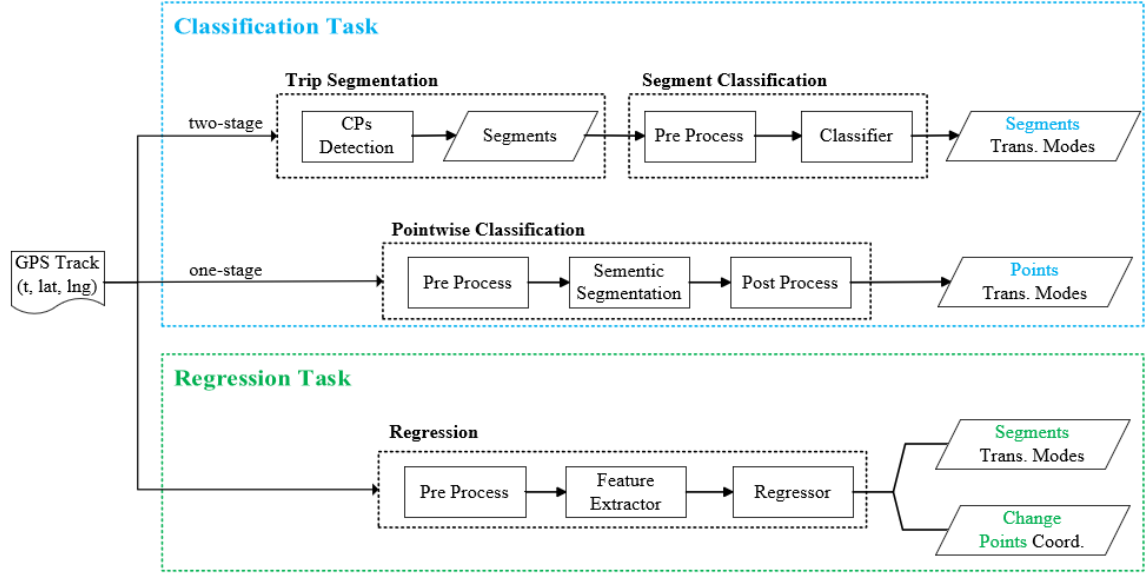

**Figure 3 The proposed regression task and existing classification task. CP: change point.**

**RQ1**: *how to reframe change point detection as supervised regression problem*?

The key is to define coordinate for change point as done in YOLO and SSD. However, one can takes one and only one unique transportation mode at a moment, so, we can't regress bounding box size to cover the segment as proper as possible as in object detection (*8-9*). To tackle this, we alternatively choose to regress coordinates or locations (**Equation 4**) of change points:

$$l = \frac{N_{CP}}{N} \in (0, 1] \qquad (4)$$

where $N_{CP}$ denotes the number of GPS points previous the change point and $N$ denotes the length of the trip. For example, the coordinates of the two change points in **Figure 2** is 9/(9+9+6)=0.375 and (9+9)/(9+9+6)=0.75 respectively which are the ground truths providing supervisory information for regression.
9

**RQ2**: *how to unify change point detection and segment classification?*

Noting that both coordinate and class probability range from 0 to 1, so, it's reasonable to put them together as outputs for a regression problem. Considering the non-linearity between coordinate/probability and trajectory, the traditional linear regression is incompetent and thus the deep learning methods are chosen. Sigmoid (**Equation 5**) is taken as activation function of output layer.

$$Sigmoid(x) = \frac{1}{1 + e^{-x}} \tag{5}$$

**RQ3**: *what's the optimization objective?*

We optimize the weighted sum of localization loss and classification loss for our model as **Equation 6**:

$$L(l, \hat{l}, p, \hat{p}) = \lambda_{\text{loc}} \underbrace{\sum_{i=1}^{n_{\text{CP}}} (l_i - \hat{l}_i)^2}_{\text{Loc Loss}} + \lambda_{\text{cls}} \underbrace{\sum_{i=1}^{n_{\text{CP}}+1} N_i^{\text{seg}} \sum_{j=1}^{k} (p_{ij} - \hat{p}_{ij})^2}_{\text{Cls Loss}} \tag{6}$$

where $l, \hat{l}$ denotes the ground truth and predicted value of location of the change point, $p, \hat{p}$ denotes the ground truth and predicted value of the transportation mode, $\lambda_{\text{loc}}$ and $\lambda_{\text{cls}}$ is positive and used for balancing localization and classification loss. $n_{\text{CP}}$ is number of true change points in a trip and $k$ is number of travel modes. Both losses are selected as the sum-squared error which is easy to optimize. If a trip has no change points, i.e., $n_{\text{CP}} = 0$, then related localization loss is zero. By means of predicted change points, we can divide the trip into segments and assign them corresponding class probabilities. To reflect the various lengths of segments, we weight a segment by its length $N_i^{\text{seg}}$.

**Two Frameworks for Solving**

To solve the reframed regression problem, referring to YOLO and SSD, we design two frameworks based on deep learning, which are called TrajYOLO and TrajSSD (see **Figure 4**).

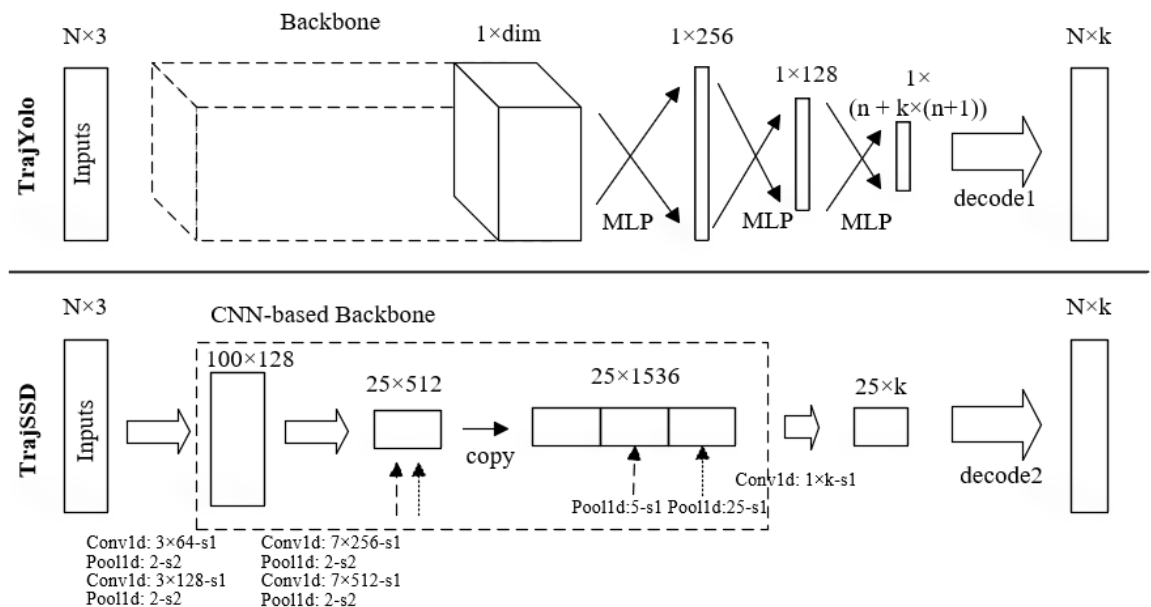

**Figure 4 Two frameworks for solving the unified regression task. TrajYOLO: features can be extracted manually or automatically and then sent to MLP to produce results. TrajSSD: features can only be extracted by CNN-based backbone and then sent to small size convolutional kernels to produce results.**

*The Framework of TrajYOLO*

Referring YOLO (*8*), the first framework is called TrajYOLO, depicted in up part of **Figure 4**. After extracting features from the trip, we flatten and feed them to MLP to generate coordinates and class probabilities. For example, we regress $n$ change points and associated $n+1$ segments across $k$ transportation modes for a trip, then there are $n + k \times (n + 1)$ outputs in total of the MLP. Under this framework, the change points are searched *directly* and features can be extracted manually or automatically. It's convenient to integrate human knowledge if we have enough understanding about GPS trajectories and human activities.

The following backbones are used for feature extraction:
- **11F**: 11 powerful hand-crafted features from Li et al. (*19*), that is, mean, variation and 85% percentile of speed, acceleration and head direction change, total distance and ratio of low-speed points.
- **MLP**: also known as Fully Connected Layers (FCL), the neuron in MLP connects to all neurons of preceding layer.
- **PointNet**: a classic network in point cloud (*24*), which was modified by the one-stage study (*7*) to extract features, which are obtained by a max-pooling layer following MLP layers.
- **CNN**: compared to MLP, the neuron in ConvNet only connects to a small set of neurons of preceding layer and connections in the same layer share weights, which reduce parameters and retain spatial information. The architecture of convolution followed by activation and pooling is adopted. Every time down sampling data by pooling of size 2, we double the channels.
- **ResNet**: the deep residual networks are variant of CNN and solve the vanishing gradient problem by short-cut link, making neural networks deep enough and extract high level features (*25*).
- **LSTM and Bi-LSTM**: the former can capture sequential dependencies from beginning to end in a sequence whereas the latter can extract features from two directions, i.e., beginning to end and the reverse direction.

For MLP, CNN, ResNet, LSTM and Bi-LSTM, the number of output channels of first layer is set as 64, and the channels doubles once the number of layers increases. That is, the output channels are 64 → 128 → 256…

*The Framework of TrajSSD*

Like in YOLO, the main weakness of TrajYOLO is inaccurate localization. To tackle this, in object detection, the practical method is to introduce anchor boxes and FCN (*9*). Anchor boxes of various predefined sizes and shapes from prior serve as reference, making it easier for bounding boxes regression. FCN retains positional information well due to convolution's intrinsic characteristics to fuse spatial information. Similarly, we predefine the potential coordinates of change points. Our practice is distributing potential change points uniformly by length of $l_{uni}$, e.g., $l_{uni} = 20$, then the $21^{th}$, $41^{th}$, $61^{th}$ … GPS points are candidates, which divide a trip into sub-trips of same length of $l_{uni}$. *Directly* picking a set of candidates is intractable, hence, we decide to select change points *indirectly*, i.e., we classify the sub-trips with uniform length first and then find change points according to the classification results. For instance, the second sub-trip and the third sub-trip are predicted as walk and bus respectively, then the second candidate is taken as a predicted change point and can engage in loss calculation. Otherwise, the second candidate is neglected.

To implement this, referring another famous work single shot multi-box detector (SSD) (*9*), we created a framework based on fully convolutional neural networks, named TrajSSD, depicted in down part of **Figure 4**. In TrajSSD, the raw input $N \times 3$ is mapped into the high dimensional space by successive convolution and pooling layers to be dense representation (e.g., 25×1536 in **Figure 3**), which is then filtered by a set of convolutional kernels of small size (e.g., 1 in **Figure 4**) to produce predictions. The outputs of $25 \times k$ represent class probabilities of 25 sub-trips across $k$ transportation modes, where 25 is changeable and determined by $l_{uni}$ and *N*. In our case, 25 = $N/l_{uni}$ = 400/16. The main advantage is that MLP is replaced by small convolutional kernels and thus positional information is retained, computational cost and training time decreases, localization accuracy increases. Nonetheless, the framework requires that CNN leads the feature extraction process which is unbeneficial to fuse human knowledge.

Visually, TrajSSD is analogous to a traditional two-stage method which divides the trip into segments using uniform size window and then classify. The significance lies in that TrajSSD processes trips instead of segments and thus utilize more context. In addition, TrajSSD involves localization error in loss, making networks converge faster.

To extract multi-scale spatial information, we take advantage of the simple but robust Pointwise Pyramid Pooling (3P) module (*10*). For outputs of the last convolutional layer, we slide pooling windows or kernels of various sizes with stride 1 over the data dimension to obtain features with different granularities. These features are then concatenated. The process can be expressed by **Equation 7**:

$$F_{3P}(P_1, \ldots, P_{N'}) = \left[ pool_{P=P_1,\ldots,P_{N'}}(f_{conv}, k_1), \ldots, pool_{P=P_1,\ldots,P_{N'}}(f_{conv}, k_q) \right] \quad (7)$$

where $k_i$ denotes one-stride pooling window size, $f_{conv}$ represents high-level features learned by convolutional layers, $N'$ is the size of $f_{conv}$ (e.g., 25 in **Figure 4**). The range $\{P = P_1, \ldots, P_{N'}\}$ out of pooling suggests that the operation is pointwise, i.e., yielding the same size of features as input vectors. As a comparison, stride of normal pooling is equal to pooling window size, thus reduce size of inputs. Small window size captures local context whereas the big one indicates global trend. Especially, $k_i = N'$ equals to global pooling. Obviously, 3P takes only non-parametric pooling operation and thus requires negligible computation.

For feature extraction, the following CNN-based backbones are adopted.
- **CNN**: the convolutional kernels of the first two layer are of size 3 to capture local details whereas sizes of the following ones are set as 7 to extract local trend.
- **3P-CNN**: we apply the 3P module to feature map of the last convolutional layer. Then, pooling results are concatenated with feature map along feature channel. Two pooling windows of sizes $k_1 = N'$ and $k_2 = 5$ are adopted to extract global and local context respectively. For instance, representation of shape 25×512 becomes 25×1536 after 3P in **Figure 4**.

**Training**

We will discuss some important issues during training.

**Data alignment**. To utilize batch gradient descent and batch normalization to speed up the training and convergence, we pad with zero trips to be length of $N_{\max}$. But the padded part will not involve in loss calculation. When inferencing, we can still receive trips of various lengths.

**Change points**. We must allocate the candidate a ground truth during training so as to calculate localization loss. After limiting length of the trip between $N_{\min}$ and $N_{\max}$, nearly all trips contain less than 3 change points. In consequence, we predict 2 change points for each trip in TrajYOLO, where the two candidates are responsible for the first and second true change points respectively. If only one true change point exists, the second candidate is overlooked and out of loss calculation. Other cases are handled similarly. For convenience, locations of change points are divided by $N_{\max}$ instead of $N$. In TrajSSD, a true change point is predicted by its nearest candidate. Candidates without matched ground truth are neglected.

**Decoding**. Obtaining outputs from TrajYOLO or TrajSSD, we need to transform them into format of $N \times k$, i.e., the class probabilities of each GPS point in a trip. For TrajYOLO, we can calculate the start and end position of each segment with the aid of coordinates of predicted change points and assign class probabilities to these segments. If the predicted coordinate is bigger than next one, all subsequent coordinates are regarded as invalid and out of loss calculation. For TrajSSD, in the final outputs ($25 \times k$ in **Figure 4**), each row represents the class probabilities of a sub-trip of the length $l_{uni}$. We assume that the whole sub-trip takes the same travel mode.

$l_{\textbf{uni}}$. The value of $l_{uni}$ determines the locations of potential change points. In fact, the $l_{uni}$ is determined by the stride of convolution kernels and the size of pooling kernels. If the strides and sizes of successive convolution and pooling kernels are $s_{conv}^1, \ldots, s_{conv}^n$ and $s_p^1, \ldots, s_p^m$ respectively, then $l_{uni} = s_{conv}^1 \times \ldots \times s_{conv}^n \times s_p^1 \times \ldots \times s_p^m$. In case of **Figure 4**, all convolutional kernels are of stride 1 and four max-pooling kernels of size 2, so $l_{uni} = 2 \times 2 \times 2 \times 2 = 16$. But the pointwise pyramid pooling has no influence on $l_{uni}$, because it doesn't reduce the size of feature mapping. For one thing, $l_{uni}$ should be small enough to grantee that each true change point has a candidate as close as possible. Inevitably, when $l_{uni} > 1$, such mechanism would misclassify some GPS points, because we can't ensure candidates cover all true change points. For another, more pooling layers can reduce computational cost, prevent overfitting and improve generalization ability. To make trade-offs, we set $l_{uni} = 16$ as default, which is near $N_{min}$ and works best in experiments.

# RESULTS
## Evaluation Criteria

The following commonly used metrics are adopted: precision, recall, accuracy and weighted F1-score (F1), calculated as **Equation 8-11** respectively.

$$Precision = \frac{TP}{TP + FP} \quad (8)$$

$$Recall = \frac{TP}{TP + FN} \quad (9)$$

$$F1 - score = \frac{2 \times Precision \times Recall}{Precision + Recall} \quad (10)$$

$$Accuracy = \frac{No.\,of\,true\,classified\,samples}{No.\,of\,all\,samples} \quad (11)$$

where TP, FP, FN denote the number of true positives, false positives and false negatives respectively. Weighted F1-score is the weighted average of F1-score across all classes. Note, a GPS point is taken as a sample, so, the accuracy by point is denoted as $A_P$. Precision, recall and F1-score measure the ability of model to distinguish one class from others, and accuracy and weighted F1-score reflect model's overall performance.

For change point detection, we take precision and recall as criteria. A true change point is considered as true positive if there is a predicted change point within 150 meters (*5*). In fact, the denominator of precision is the number of predicted change points and the dominator of recall is the number of true change points.

**Experimental Setup**

After all pre-processing steps, in total, we obtain 11845 trips of lengths between $N_{min}$ and $N_{max}$ (i.e., 20 and 400). These trips are split into training, validation and test set by ratio 7:1:2. All pooling operation is selected as max-pooling. To ameliorate overfitting, we use dropout of ratio 0.5 in the first two layers of the final MLP layers and carry out the early stopping strategy, i.e., stopping training if loss don't decrease by 0.001 over 15 epochs. ReLU is taken as activation function for all layers except for the last fully connected layer with Sigmoid. In all experiments, optimizer is Adam with initial learning rate of 0.001, which drops to 1/10 of the previous value every ten epochs util less than $10^{-7}$. The model is trained over at most 100 epochs with batch size 128. $\lambda_{loc} = 300$ and $\lambda_{cls} = 1$.

All methods are coded with Python programming language of version 3.7. The networks are implemented in Pytorch 1.6. All experiments are run on a server with a NVIDIA GeForce GTX 1080 GPU.

**Results**

The results of various backbones under two frameworks are listed in **TABLE 1**. The best network architectures, i.e., number of neural network layers, are found out by grid search and validation set. For TrajYOLO, the hand-crafted features work worst, because such features are designed for the purpose of GPS segment classification, which focus on the global trend of a GPS segment. However, under our modeling, what we receive is the trip instead of the single-one mode segment, hence requiring more details to detect the change of underlying transportation mode. The MLP and PointNet misses the local structures too much so the performance is not outstanding. The convolutional neural network outperforming others with great margin (at least 3.5% in $A_P$), we attribute this to its advantages to capture local spatial structures and fuse local information, which is fundamental for change point detection and thus for the overall identification. The ResNet, a variant of CNN, is surprisingly inferior to CNN greatly. The comparison between confusion matrices of ResNet and CNN indicates that ResNet struggles with discerning car from other transportation modes especially bus and mistaking bus as car. As we know, the motion characteristics of car and bus share many similarities (*6*), making it hard to distinguish between car and bus directly based on such features. Therefore, we need to map the motion characteristics into another feature space to achieve disentanglements (e.g., by neural networks), whereas the unique shortcut connections in ResNet preserve the raw features to some extent, making differentiating between car and bus harder. The results of LSTM and Bi-LSTM demonstrate that time dependencies are not so important for this task. Nevertheless, all backbones under TrajYOLO struggle with change point detection because of loss of spatial information resulted from the flatten operation in the last fully connected layers.

For TrajSSD, two backbones have excellent performances. With the nearly same architectures, CNN under TrajSSD found much more change points than that of CNN under TrajYOLO, i.e., recall rate increasing from 10.0% to 60.9%, whereas the latter performs batter in terms of the overall transportation mode identification. The reason lies in that small size convolutional kernels in TrajSSD replace MLP in TrajYOLO, retaining positional information without of loss and making it easier to detect change points, however, small size convolutions lack of ability to seize general trend, making it harder to identify transportation modes. To enhance the ability to capture information with various granularities, we add the 3P structure at the end of CNN. The results show that the 3P plays a huge role, that is, precision of change point detection increases from 14.8% to 29.2%, $A_P$ from 81.9% to 85.3% and F1 from 81.7% to 85.2%, whereas recall rate of change point detection decreases from 60.9% to 53.5%. We believe such price is worth especially considering that 3P brings little extra computational cost.

The confusion matrix of the best result (3P-CNN under TrajSSD) is detailed in **TABLE 2**. From F1-score, all five transportation modes can be distinguished well and the identification performances in order are: train > bike > walk > bus = car. Among them, train is the easiest to identify with the highest recall of 97.0% in literatures due to its simple and special motion pattern, i.e., uniform linear motion. In addition, 3P-CNN can distinguish between car and bus much better than others, making it work best, which is ascribed to its ability to capture their local differences. The bottleneck lies in confusion between car and bus and mistaking bike and bus as walk. Similar motion patterns of car and bus make them indistinguishable. Slow speed of bike and many stops of bus make them resemble walk. In the future, more techniques and tricks can be exploited to extract finer details.

**TABLE 1 Results of various backbones under two frameworks. Bold value indicates maximum of a column under TrajYOLO and TrajSSD. The selected best backbone is denoted with asterisk.**

| Framework | Backbone | Optimal No. of layers | Change point detection | | Transportation mode identification | |
|---|---|---|---|---|---|---|
| | | | Recall | Precision | $A_P$ | F1 |
| TrajYOLO | 11F | - | 0.025 | 0.063 | 0.651 | 0.644 |
| | MLP | 3 | 0.068 | 0.197 | 0.774 | 0.773 |
| | PointNet | - | 0.009 | 0.125 | 0.793 | 0.793 |
| | CNN* | 5 | 0.100 | 0.249 | **0.834** | **0.834** |
| | ResNet | 5 | **0.104** | 0.071 | 0.742 | 0.741 |
| | LSTM | 5 | 0.070 | 0.184 | 0.799 | 0.798 |
| | Bi-LSTM | 4 | 0.101 | **0.285** | 0.796 | 0.795 |
| TrajSSD | CNN | 4 | **0.609** | 0.148 | 0.819 | 0.817 |
| | 3P-CNN* | 4 | 0.535 | **0.292** | **0.853** | **0.852** |

TABLE 2 Confusion matrix of the best result (3P-CNN under TrajSSD). $A_P$=0.853.

| PointNet | | Prediction | | | | | Recall |
|---|---|---|---|---|---|---|---|
| | | Walk | Bike | Bus | Car | Train | |
| Ground Truth | Walk | **232133** | 6590 | 13245 | 3637 | 1430 | 0.903 |
| | Bike | 15480 | **135651** | 5445 | 553 | 349 | 0.861 |
| | Bus | 29758 | 3557 | **182237** | 21094 | 3646 | 0.758 |
| | Car | 8911 | 1001 | 14163 | **109522** | 3913 | 0.796 |
| | Train | 1302 | 191 | 748 | 1918 | **136075** | 0.970 |
| Precision | | 0.807 | 0.923 | 0.844 | 0.801 | 0.936 | - |
| F1-score | | 0.852 | 0.891 | 0.799 | 0.799 | 0.953 | - |

**Comparison With Related Studies**

In this sub-section, we compare our best model (3P-CNN under TrajSSD) with related works on the same test set in terms of change point detection and transportation mode identification. All repeated works are implemented in Python and by various packages, e.g., ruptures for PELT, scikit-learn for classic machine learning algorithms and pytorch for neural networks. Related codes are also available in our repository.

For change point detection, the following four methods are implemented: uniform time window segment (UTW) (*5*), walk segment (WS) (*5*), time window and similarity (TWS) (*15*) and PELT (*16*). As shown in **Figure 5**, our method struggles with recall but successes in precision. This is because existing methods prefer over segmentation strategy and thus search much more candidate change points than true change points, whereas candidates in our method are much fewer.

Li et al. (*7*) have conducted exhaustive experiments to find the best match between change point detection algorithms and segment classification algorithms of two-stage methods so as to compare the overall identification performances with their one-stage method, i.e., inferring transportation modes from raw unlabeled GPS trajectories. Following this, we repeated the representative machine learning and deep learning methods and compared them with ours on the same training and test set. For two-stage methods, training and test trips are divided into segments by the same segmentation method; inputs of traditional machine learning methods are 11 powerful hand-crafted features from Li et al. (*19*) as mentioned above. Usually, $A_P$ and weighted F1-score have little difference so we just report the former. As can be seen from **TABLE 3**, our method can achieve competitive result. Another, 3P-CNN under TrajSSD is light-weighted with model size of just 4.76 MB and the training is fast, i.e., converging in 40 epochs in less than 5 minutes, making it convenient to retrain in new dataset and deploy in real-application.

Clearly, our models and the one-stage method (*7*) are ahead of others because both take as inputs trips instead of segments so that leveraging more context. The evidence is that average lengths of segments obtained by UTW, WS, TWS and PELT are 48, 112, 126 and 34 respectively, whereas that of trips in ours and the one-stage study is 310. Another is that with the nearly same inputs and architectures, CNN under TrajYOLO performs much better than under two-stage framework ($A_P$ 0.834 vs. 0.782).

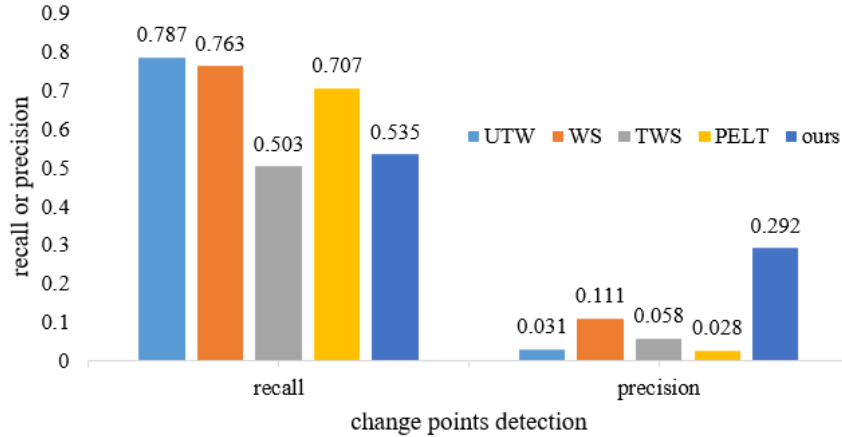

**Figure 5 Comparison with related works in terms of change point detection.**

**TABLE 3 Comparison with related studies in terms of overall identification performance. CPD: change point detection, TMI: transportation mode identification, WS: walk segment, TWS: time window and similarity.**

| Task | Framework | CPD algorithm | Classification algorithm | $A_P$ |
|---|---|---|---|---|
| Classification | Two-stage | WS | KNN | 0.731 |
| | | WS | SVM | 0.641 |
| | | WS | MLP | 0.758 |
| | | WS | DT | 0.756 |
| | | WS | RF (Li et al, 2021) | 0.792 |
| | | TWS | CNN (Dabiri and Heaslip, 2018) | 0.782 |
| | | TWS | CNN ensemble (Dabiri and Heaslip, 2018) | 0.804 |
| | | TWS | LSTM (Yu, 2019) | 0.793 |
| | One-stage | BBN (Feng and Timmermans, 2013) | | 0.626 |
| | | RF (Prelipcean et al., 2014) | | 0.629 |
| | | Bi-GRU + maxout (Jiang et al., 2017) | | 0.791 |
| | | 3P-MSPointNet (Li et al, 2023) | | 0.849 |
| Regression | TrajYOLO | CNN (ours) | | 0.834 |
| | TrajSSD | 3P-CNN (ours) | | **0.853** |

**Ablation Studies**

We ablate our best TrajYOLO (CNN) and TrajSSD (3P-CNN) using the default settings in **TABLE 4**.

For TrajYOLO, the number of candidate change points $\hat{n}_{CP}$ has little effect when $\hat{n}_{CP} > 1$. Because nearly all trips contain less than 3 change points and a true change point is only predicted by a candidate, more candidates make no sense. Weight of localization loss $\lambda_{loc}$ also make little difference because the classification of each GPS point implies change point detection to some extent. For TrajSSD, we alter the pooling window size of first pooling layer as 1, 3 and 4 so as to adjust the length of sub-trip $l_{uni}$ to 8, 24 and 32 respectively. When $l_{uni} = 32$, the performance drops drastically whereas $l_{uni}$ = 8, 16 and 24 work well. Usually, we can take a sub-trip of length less than 20 (*5*) as single-one mode segment, however, the sub-trip of length of 32 has higher possibility to involve more than one transportation modes. From **TABLE 4d**, the size of small convolutional kernels functioning to replace MLP has linear effect, i.e., the bigger $k_s$, the worse performance. This is because MLP is in fact convolution of kernel size same as input size, which fuses global information and losses local details. In consequence, smaller $k_s$ is more helpful to preserve local spatial structures to increase localization accuracy.

**TABLE 4 Ablation experiments. Maximum of each column is in bold. Default setting is marked in gray. CPD: change point detection; TMI: transportation mode identification.**

| $\hat{n}_{CP}$ | CPD Recall | Precision | TMI $A_P$ | F1 |
|---|---|---|---|---|
| 1 | 0.030 | 0.120 | 0.824 | 0.824 |
| 2 | 0.100 | **0.249** | **0.834** | **0.834** |
| 3 | 0.077 | 0.191 | 0.828 | 0.827 |
| 4 | **0.101** | 0.213 | 0.829 | 0.829 |

(a) Number of change point candidates.

| $\lambda_{loc}$ | CPD Recall | Precision | TMI $A_P$ | F1 |
|---|---|---|---|---|
| 200 | 0.086 | 0.228 | 0.833 | 0.834 |
| 300 | **0.100** | **0.249** | **0.834** | **0.834** |
| 400 | 0.083 | 0.191 | 0.832 | 0.832 |
| 500 | 0.091 | 0.218 | 0.831 | 0.831 |

(b) Weight of localization loss.

| $l_{uni}$ | CPD Recall | Precision | TMI $A_P$ | F1 |
|---|---|---|---|---|
| 8 | **0.567** | 0.229 | 0.848 | 0.847 |
| 16 | 0.535 | 0.292 | **0.853** | **0.852** |
| 24 | 0.442 | **0.315** | 0.850 | 0.849 |
| 32 | 0.067 | 0.030 | 0.812 | 0.812 |

(c) Length of sub-trip.

| $k_s$ | CPD Recall | Precision | TMI $A_P$ | F1 |
|---|---|---|---|---|
| 1 | **0.535** | 0.292 | **0.853** | **0.852** |
| 3 | 0.493 | 0.300 | 0.850 | 0.849 |
| 5 | 0.500 | 0.307 | 0.847 | 0.846 |
| 7 | 0.481 | **0.316** | 0.842 | 0.841 |

(d) Size of small convolutional kernels.

**CONCLUSION**

In this paper, we propose to reframe the transportation mode identification based on GPS trajectories as a regression instead of the previous classification task. Under the new task, the change point detection and segment classification are unified in a single regression problem, so that change points can be detected in supervised manner and the whole identification process can be optimized conveniently. To solve the regression problem, two frameworks are proposed. TrajYOLO receives extracted features and directly predict the coordinates of change points and class possibilities of associated segments. TrajSSD is based on fully convolutional networks and replace fully connected layers with small size convolutional kernels, thus retaining spatial information and increase recall of change points. We exploited various backbones to extract features, including hand-crafted, MLP, CNN, ResNet and LSTM. Among them, CNN works best due to its ability to represent features and hold local spatial information, which is enhanced again by pointwise pyramid pooling (3P) module. Exhaustive studies were conducted to validate the proposed method and search the optimal parameters. Comparison with related studies show that our method struggles with recall but successes in precision in terms of change point detection, and has obvious advantages over most of existed works in terms of the overall identification performance. We attribute these to our supervised change point detection and utilization of longer context.

For future work, it's promising to incorporate priors to change point detection and design finer network architectures to capture more details. In addition, one can try to predict locations of change points and class of segments with two independent sub-networks.

1
2